%% file: main.tex
\documentclass[11pt]{article}
\usepackage{nips15submit_e,times}
\usepackage{hyperref}
\usepackage{url}

\usepackage[numbers,sort&compress]{natbib}

\usepackage{algorithm}
\usepackage{algorithmic}

\usepackage{graphicx}
\usepackage{comment}
\usepackage{amsmath}
\usepackage{amssymb}
\usepackage{siunitx}
\usepackage{caption}
\usepackage{subcaption}
\usepackage{wrapfig}
\usepackage{dsfont}
\usepackage{siunitx}
\input{include/murphy}

\DeclareMathOperator*{\argmin}{arg\,min}

\newcommand{\figspace}{\vspace{-2mm}}

\title{Data-Efficient Learning of Feedback Policies from Image Pixels using Deep  Dynamical Models}
\author{
John-Alexander M. Assael\thanks{These authors contributed equally to this work.}\\
Department of Computing \\
Imperial College London, UK\\
\texttt{i.assael@imperial.ac.uk}
\And
Niklas Wahlstr{\"o}m\footnotemark[1]\\
Division of Automatic Control \\
Link\"oping University, Sweden\\
\texttt{nikwa@isy.liu.se}
\And%\AND
Thomas B. Sch\"on \\
Department of Information Technology\\
Uppsala University, Sweden\\
\texttt{thomas.schon@it.uu.se}
\And
Marc Peter Deisenroth \\
Department of Computing \\
Imperial College London, UK\\
\texttt{m.deisenroth@imperial.ac.uk}
}

\nipsfinalcopy % Uncomment for camera-ready version

\newcommand{\myargmin}[2]{\underset{#1}{\mathrm{arg}\,\mathrm{min}}\,\,#2}

\newcommand{\tind}{t}
\newcommand{\tend}{N}

\begin{document}

\maketitle
\begin{abstract}
Data-efficient reinforcement learning (RL) in continuous state-action spaces using very high-dimensional observations remains a key challenge in developing fully autonomous systems. We consider a particularly important instance of this challenge, the pixels-to-torques problem, where an RL agent learns a closed-loop control policy (``torques'') from pixel information only. We introduce a data-efficient, model-based reinforcement learning algorithm that learns such a closed-loop policy directly from pixel information. The key ingredient is a deep dynamical model for learning a low-dimensional feature embedding of images jointly with a predictive model in this low-dimensional feature space. Joint learning is crucial for long-term predictions, which lie at the core of the adaptive nonlinear model predictive control strategy that we use for closed-loop control. Compared to state-of-the-art RL methods for continuous states and actions, our approach learns quickly, scales to high-dimensional state spaces, is lightweight and an important step toward fully autonomous end-to-end learning from pixels to torques. 
\end{abstract}

\section{Introduction}
The vision of fully autonomous and intelligent systems that learn by themselves has inspired artificial intelligence~(AI) and robotics research for many decades.
% pixels-to-torques problem
The \emph{pixels to torques problem} identifies key aspects of such an autonomous system: autonomous thinking and decision making using (general-purpose) sensor measurements only, intelligent exploration and learning from mistakes.  We consider the problem of efficiently learning closed-loop policies (``torques'') from pixel information end-to-end. Although, this problem falls into the general class of reinforcement learning (RL)~\cite{sutton1998reinforcement}, it is challenging for the following reasons: (1) The state-space (here defined by pixel values) is enormous (e.g., for a $50 \times 50$ image, we are looking at $2500$ continuous-valued dimensions); (2) In many practical applications, we need to find solutions data efficiently: When working with real systems, e.g., robots, we cannot perform millions of experiments because of time and hardware constraints.

% learn forward models: data efficiency
One way of using data efficiently, and, therefore, reducing the number of experiments, is to learn predictive forward models of the underlying dynamical system, which are then used for internal simulations and policy learning. These ideas have been successfully applied to RL, control and  robotics~\cite{Schmidhuber1990,atkeson1997learning,bagnell2001autonomous,DeisenrothFR:2015,pan2014probabilistic,levine2015end}, for instance. However, they often rely on heuristics, demonstrations or engineered low-dimensional features, and do not easily scale to data-efficient RL using pixel information only.

% dimensionality reduction
A common way of dealing with high-dimensional data is to learn low-dimensional feature representations. Deep learning architectures, such as deep neural networks~\cite{hinton2006reducing}, stacked auto-encoders~\cite{bengio2007greedy,vincent2008extracting}, or convolutional neural networks~\cite{lecun1998gradient}, are the current state-of-the-art in learning parsimonious representations of high-dimensional data. 
Since 2006, deep learning has produced outstanding empirical results in image, text and audio tasks~\cite{schmidhuber2014deep}.

% some related work
\paragraph{Related Work} 
Over the last months, there has been significant progress in the context of the pixels-to-torques problem. 
% Mnih DQN
A first working solution was presented in 2015~\cite{mnih2015human}, where an RL agent automatically learned to play Atari games purely based on pixel information. The key idea was to embed the high-dimensional pixel space into a lower-dimensional space using deep neural networks and apply Q-learning in this compact feature space. A potential issue with this approach is that it is not a data-efficient way of learning policies (weeks of training data are required), i.e., it will be impractical to apply it to a robotic scenario. This data inefficiency is not specific to Q-learning, but a general problem of model-free RL methods~\cite{atkeson1997learning,Schneider1997}.

% Wahlstrom & Watter
To increase data efficiency, model-based RL methods aim to learn a model of the transition dynamics of the system/robot and subsequently use this model as a surrogate simulator. Recently, the idea of learning predictive models from raw images where only pixel information is available was exploited~\cite{wahlstrom2015pixels,watter2015e2c}. The approach taken here follows the idea of Deep Dynamical Models (DDMs)~\cite{wahlstrom2015learning}: Instead of learning predictive models for images directly, a detour via a low-dimensional feature space is taken by embedding images into a lower-dimensional feature space, e.g., with a deep auto-encoder. This detour is promising since direct mappings between high-dimensional spaces require large data sets. Whereas \citet{wahlstrom2015pixels} consider deterministic systems and nonlinear model predictive control (NMPC) techniques for online control, \citet{watter2015e2c} use variational auto-encoders~\cite{JimenezRezende2014a}, local linearization, and locally linear control methods (iLQR~\cite{Todorov2005} and AICO~\cite{Toussaint2009}).

% \todo[inline]{Say something about the huge number of parameters that need to be learned. Niklas is concatenating features. I couldn't see what Watter et al. do. Might have missed it.} 

To model the dynamical behavior of the system, the pixels of both the current and previous frame are used. \citet{watter2015e2c} concatenate the input pixels to discover such features, whereas \citet{wahlstrom2015pixels} concatenate the processed low-dimensional embeddings of the two states. The latter approach requires at least ${\approx}4\times$ fewer parameters, which makes it a promising candidate for  more data-efficient learning.
Nevertheless, properties such as local linearization  \cite{watter2015e2c} can be attractive. However, the complex architecture proposed by \citet{watter2015e2c} is based on very large neural network models with ${\approx}6$ million parameters for learning to swing up a single pendulum.
A vast number of training parameters results in higher model complexity, and, thus, decreased statistical efficiency. Hence, an excessive number of training samples, which might not be available, is required to learn the underlying system, taking several days to be trained.
These properties make data-efficient learning complicated. Therefore, we propose a relatively lightweight architecture to address the pixels-to-torques problem in a data-efficient manner.

% state what we do differently: contribution
\paragraph{Contribution}
We propose a data-efficient model-based RL algorithm that addresses the pixels-to-torques problem. (1) We devise a data-efficient policy learning framework  based on the DDM approach for learning predictive models for images. We use state-of-the-art optimization techniques for training the DDM. (2) Our model profits from a concatenation of low-dimensional features (instead of high-dimensional images) to model dynamical behavior, yielding ${\approx}4$--$20$ times fewer model parameters and faster training time than the complex E2C architecture~\cite{watter2015e2c}. 
In practice, our model requires only a few hours of training, while E2C~\cite{watter2015e2c} requires days.
(3) We introduce a novel training objective that encourages consistency in the latent space paving the way towards more accurate long-term predictions. 
Overall, we use an efficient model architecture, which can learn tasks of  complex non-linear dynamics.

\section{Problem Set-up and Objective}

We consider an $N$-step finite-horizon RL setting in which an agent attempts to solve a particular task by trial and error. 
In particular, our objective is to find a closed-loop policy $\pi^*$, that minimizes the long-term cost $V^\pi = \sum_{\tind=0}^{\tend-1} c(\vs_\tind, \vu_\tind)$, where $c$ denotes an immediate cost, $\vs_\tind\in\real^{N_s}$ is the continuous-valued system state, and $\vu_\tind\in\real^{N_u}$ are continuous control signals. 

% challenges
The learning agent faces the following additional two challenges: (a) The agent has no access to the true state $\vs_t$, but perceives the environment only through high-dimensional pixel information $\vx_\tind\in\real^{N_x}$ (images); (b) A good control policy is required in only a few trials. This setting is practically relevant, e.g., when the agent is a robot that is monitored by a video camera based on which the robot has to learn to solve tasks fully autonomously. 
% The agent perceives that a task is completed when a desired reference video frame is obtained. 
Therefore, this setting is an instance of the pixels-to-torques problem.

We solve this problem in three key steps, which will are detailed in the following sections: 
(a) Using a deep auto-encoder architecture we map the high-dimensional pixel information $\vx_\tind$ to a low-dimensional embedding/feature $\vz_\tind$.
(b) We combine the $\vz_\tind$, and $\vz_{\tind-1}$ features with the control signal $\vu_\tind$ to learn a predictive model of the system dynamics for predicting future features $\vz_{\tind+1}$. (a) and (b) form a Deep Dynamical Model (DDM)~\cite{wahlstrom2015learning}.
(c) We apply an adaptive nonlinear model predictive control strategy for optimal closed-loop control and end-to-end learning from pixels to torques.
%\todo[inline]{We propose to solve this problem in the following steps: 1, 2, 3, 4. List them all (1 sentence) and be as precise as you can in 1 sentence. The next sections detail these key steps.}

\section{Learning a Deep Dynamical Model (DDM)}
\begin{wrapfigure}{R}{0.45\textwidth}
    \centering
    \vspace{-2mm}
    \includegraphics[width=0.9\linewidth]{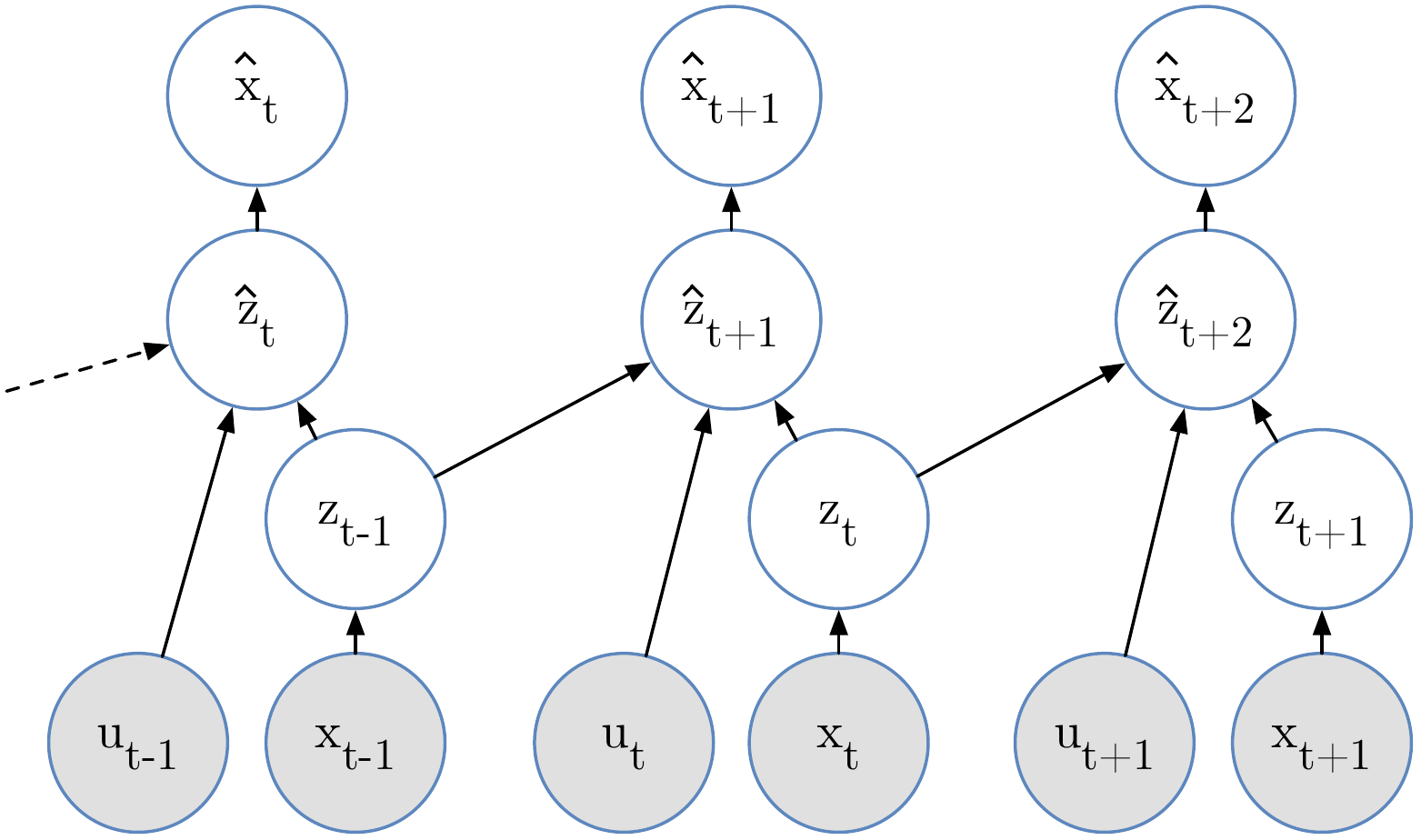}
    \caption{Graphical model of the DDM process for predicting future states. %, where pixel observations $\vx_t$ are embedded to $\vz_t$. Combining two consecutive $\vz_{t-1}$ and $\vz_t$ with the control signal $\vu_t$, we predict the next feature $\hat{\vz}_{t+1}$, from which we can generate the next predicted image $\hat{\vx}_{t+1}$.
    }
    \label{fig:graphical}
    \figspace
\end{wrapfigure}

Our approach to solving the pixels-to-torques problem is based on a deep dynamical model (DDM), see Figure~\ref{fig:graphical}, which jointly (a) embeds high-dimensional images in a low-dimensional feature space via deep auto-encoders, and (b) learns a predictive forward model in this feature space, based on the work by \citet{wahlstrom2015learning}.
In particular, we consider a DDM with control signals $\vu_\tind$ and high-dimensional observations $\vx_\tind$ at time-step $\tind$. We assume that the relevant properties of $\vx_\tind$ can be compactly represented by a feature variable $\vz_\tind$. Furthermore, $\tilde \vx_\tind$ is the reconstructed high-dimensional measurement. The two components of the DDM, i.e., the low-dimensional feature and the predictive model, which predicts future features $\hat \vz_{\tind+1}$ and observations $\hat \vx_{\tind+1}$ based on past observations and control signals, are detailed in the following sections.  
% quick notation intro
% Throughout this paper, $\vx_\tind$ denotes the high-dimensional measurements, $\vz_\tind$ the corresponding low-dimensional encoded features and $\tilde \vx_\tind$ the reconstructed high-dimensional measurement. Furthermore, $\tilde \vx_\tind$ is the reconstructed high-dimensional measurement, $\hat \vz_{\tind+1}$ and $\hat \vx_{\tind+1}$ denote a predicted feature and measurement at time $t+1$, respectively.
%\todo[inline]{Please describe the figure.}
%

\subsection{Predictive Forward Model}

Inspired by the concept of static auto-encoders \cite{rumelhart1986learning,bengio2009learning}, we turn them into a dynamical model that can predict future features $\hat{\vz}_{t+1}$ and images $\hat{\vx}_{t+1}$. Our DDM consists of the following elements:
\begin{enumerate}\itemsep1pt \parskip0pt \parsep0pt
    \item An encoder $f_{\text{enc}}$ mapping high-dimensional observations $\vx_t$ onto low-dimensional features~$\vz_t$,
    \item A decoder $f_{\text{dec}}$ mapping low-dimensional features $\vz_t$ back to high-dimensional observations $\hat \vx_t$, and 
    \item The predictive model $f_{\text{pred}}$, which takes $\vz_t,\vz_{t-1},\vu_t$ as input and predicts the next latent feature $\hat{\vz}_{t+1}$.
\end{enumerate}
The $f_{\text{enc}},f_{\text{dec}}$ and $f_{\text{pred}}$ functions of our DDM are neural network models performing the following transformations:
\begin{subequations}  \label{eq:transformations}
\begin{align}
    \vz_t &= f_{\text{enc}}(\vx_{t}), \label{eq:enc} \\
    \tilde{\vx}_t &= f_{\text{dec}}(\vz_t), \label{eq:rec} \\
    \hat{\vz}_{t+1} &= f_{\text{pred}}(\vz_{t-1},\vz_t,\vu_t), \label{eq:pred} \\
    \hat{\vx}_{t+1} &= f_{\text{dec}}(\hat{\vz}_{t+1}). \label{eq:dec}
    \end{align}
\end{subequations}
%\todo[inline]{We have a notation clash where $\hat{\vx}_t$ can be both a reconstruction and a prediction. For example $\hat{\vx}_t$ in Figure 1 and in eq (1) are not the same object. Or will this be evident from context?\\
%\bf{In the context $\hat{\vx}_t$ is the reconstruction and $\hat{\vx}_{t+1}$ the prediction. Is there a standard for denoting reconstruction?}}
%

\begin{figure}[tb]
    \centering
    \includegraphics[width=0.55\linewidth]{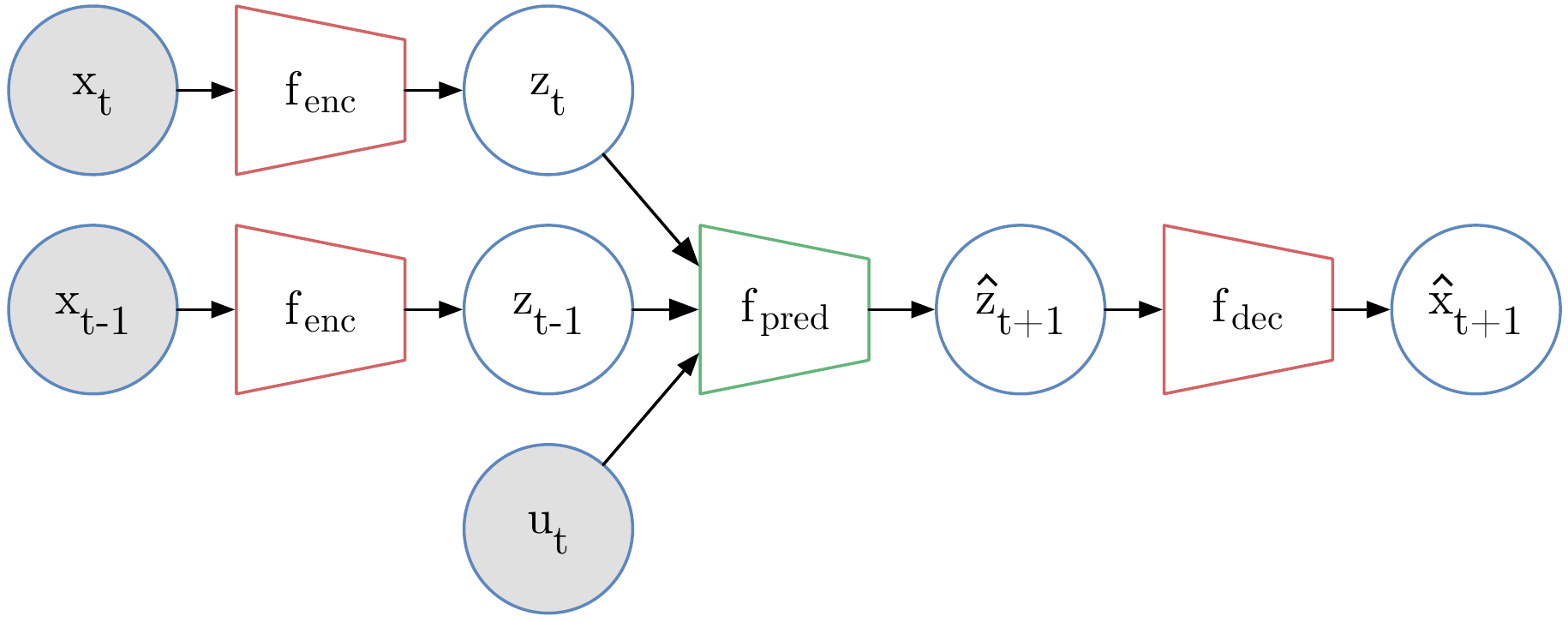}
    \caption{The architecture of the proposed DDM, for extracting the underlying properties of the dynamical system and predicting the next image frame from raw pixel information.}
    \label{fig:model_net}
\end{figure}

We now put these elements together to construct the DDM. The DDM architecture takes the raw images $\vx_{t-1}$ and $\vx_t$ as input and maps them to their low-dimensional features $\vz_{t-1}$ and $\vz_t$ respectively, using $f_{\text{enc}}$ in \eqref{eq:enc}. These latent features are then concatenated and, together with the control signal $\vu_t$, used to predict $\hat \vz_{t+1}$ with $f_{\text{pred}}$ in \eqref{eq:pred}. Finally, the predicted feature $\hat \vz_{t+1}$ is passed through the decoder network $f_{\text{dec}}$, to compute the predicted image $\hat \vx_{t+1}$. The overall architecture is depicted in Figure~\ref{fig:model_net}.

The neural networks $f_{\text{enc}},f_{\text{dec}}$ and $f_{\text{pred}}$ that compose the DDM, are parameterized by $\theta_{\text{enc}},\theta_{\text{dec}}$ and $\theta_{\text{pred}}$ respectively. These parameters consist of the weights that perform linear transformations of the input data in each neuron.

\subsection{Training}

For training the DDM in~\eqref{eq:transformations}, we introduce a novel training objective, that encourages consistency in the latent space, paving the way toward accurate long-term predictions.
More specifically, for our training objective we define the following cost functions
\begin{subequations}
\begin{align}
    \loss_{\text{R}}(\vx_t) &= \|\tilde{\vx}_t - \vx_t \|^2,\label{eq:loss1}\\
    \loss_{\text{P}}(\vx_{t-1},\vx_{t}, \vu_t,\vx_{t+1}) &= \|\hat{\vx}_{t+1} - \vx_{t+1} \|^2,\label{eq:loss2}\\
    \loss_{\text{L}}(\vz_{t-1},\vz_{t}, \vu_t,\vz_{t+1}) &= \| \hat{\vz}_{t+1} - \vz_{t+1}\|^2,\label{eq:loss3}
\end{align}
\end{subequations}
where $\loss_{\text{R}}$ is the squared deep auto-encoder reconstruction error and $\loss_{\text{P}}$ is the squared prediction error, both operating in image space. Note that $\hat{\vx}_{t+1} = f_{\text{dec}}(f_\text{pred}(f_{\text{enc}}(\vx_t), f_\text{enc}(\vx_t), \vu_t))$ depends on the parameters of the decoder, the predictive model and the encoder.
Additionally, we introduce $\loss_{\text{L}}$ that enforces consistency between the latent spaces of the encoder $f_{\text{enc}}$ and the prediction model $f_{\text{pred}}$. In the big-data regime, this additional penalty in latent space is not necessary, but if not much data is available, this additional term  increases the data efficiency as the prediction model is forced to make predictions $\hat{\vz}_{t+1} = f_{\text{pred}}(\vz_{t-1},\vz_{t}, \vu_t)$ close to the next embedded feature $\vz_{t+1}=f_{\text{enc}}(\vx_{t+1})$. The overall training objective of the current dataset $\data=(\vx_{0:N},\vu_{0:N})$ 
is given by
\begin{equation}
    \begin{aligned}
    \loss(\data) &= \sum\nolimits_{t=0}^{N-1} \loss_{\text{R}}(\vx_t) + \loss_{\text{P}}(\vx_{t-1},\vx_{t}, \vu_t,\vx_{t+1}) + \alpha \loss_{\text{L}}(\vz_{t-1},\vz_{t}, \vu_t,\vz_{t+1})\\
    &= \sum\nolimits_{t=0}^{N-1} \|\tilde{\vx}_t - \vx_t \|^2 + \|\hat{\vx}_{t+1} - \vx_{t+1} \|^2+ \alpha \| \hat{\vz}_{t+1} - \vz_{t+1}\|^2,
    \end{aligned}
    \label{eq:loss_all}
\end{equation}
where $\alpha$ is a parameter that controls the influence of $\loss_{\text{L}}$.
Finally, we train the DDM parameters $(\theta_{\text{enc}}, \theta_{\text{dec}}, \theta_{\text{pred}})$ by jointly minimizing the overall cost
\begin{equation}
    (\hat \theta_{\text{enc}}, \hat \theta_{\text{dec}},\hat \theta_{\text{pred}}) = \argmin_{\theta_{\text{enc}}, \theta_{\text{dec}}, \theta_{\text{pred}}} \loss(\data).
    \label{eq:loss_min}
\end{equation}
Training jointly leads to good predictions as it facilitates the extraction and separation of the features describing the underlying dynamical system, and not only features for creating good reconstructions~\cite{wahlstrom2015learning}.

\subsection{Network Architecture}

The neural networks $f_{\text{enc}},f_{\text{dec}}$ and $f_{\text{pred}}$ are composed by $3$ linear layers, where each of the first $2$ are followed by Rectified Linear Unit (ReLU) activation functions \cite{nair2010rectified}. As it has been demonstrated~\cite{krizhevsky2012imagenet}, ReLU non-linearities allow the network to train $\approx 6\times$ faster than the conventional $\tanh$ units, as evaluated on the CIFAR-10~\cite{krizhevsky2009learning} dataset.
Furthermore, similar to \citet{watter2015e2c}, we use Adam~\cite{kingma2014adam} to train the DDM, which is considered the state-of-the-art among the latest methods for stochastic gradient optimization.
Finally, after evaluating different weight optimization methods, such as uniform and random Gaussian~\cite{glorot2010understanding,krizhevsky2012imagenet}, the weights of the DDM were initialized using orthogonal weight initialization~\cite{saxe2013exact}, which demonstrated the most efficient training performance, leading to decoupled weights that evolve independently of each other.

\section{Policy Learning}
Our objective is to control the system to a state where a certain target frame $\vx_\text{ref}$ without any prior knowledge of the system at hand. To accomplish this
we use the DDM for learning a closed-loop policy by means of nonlinear model predictive control (NMPC). 

\subsection{NMPC using the DDM}
NMPC finds an optimal sequence of control signals that minimizes a $K$-step loss function, where $K$ is typically smaller than the full horizon. We choose to do the control in the low-dimensional embedded space to reduce the complexity of the control problem. 

Our NMPC formulation  relies on (a) a target feature $\vz_{\text{ref}}$ and (b) the DDM that allows us to predict future features. The target feature is computed by encoding the target frame $\vz_{\text{ref}} = f_{\text{enc}}(\vx_{\text{ref}}, \theta_{\text{E}})$ provided by the model. Further, with the DDM, future features $\hat \vz_1,\dotsc,\hat \vz_K$ can be predicted based on a sequence of future (and yet unknown) controls $\vu_0,\dotsc, \vu_{K-1}$ and two initial encoded features $\vz_{-1}, \vz_{0}$ assuming that the current feature is denoted by $\vz_0$.

Using the dynamical model, NMPC determines an optimal (open-loop) control sequence $\vu_0^*,\dotsc, \vu_{K-1}^*$, such that the predicted features $\vz_1,\dotsc,\hat  \vz_K$ gets as close to the target feature $\vz_\text{ref}$ as possible, which results in the objective
\begin{equation}
 \vu_0^*,\dotsc, \vu_{K-1}^* \in  \myargmin{ \vu_{0:K-1}}{\hspace{-0.5mm} \sum_{\tind = 0}^{K-1} \|\hat\vz_\tind- \vz_{\text{ref}}\|^2 \hspace{-0.5mm}+\hspace{-0.5mm} \lambda \|\vu_\tind\|^2},
 \label{eq:MPC_cost_functionc}
\end{equation}
where $\|\hat \vz_\tind - \vz_{\text{ref}}\|^2$ is a cost associated with the deviation of the predicted features $\hat \vz_{0:K-1}$ from the reference feature $\vz_\text{ref}$, and $\|\vu_\tind\|^2$ penalizes the amplitude of the control signals. Here, $\lambda$ is a tuning parameter adjusting the importance of these two objectives.
%\footnote{In the more generic description of NMPC we can straightforwardly include constraints on the states and controls in~\eqref{eq:MPC_cost_functionb}. More general cost functions or an addition penalty term that penalizes the final state $\hat x_{\tind+\pend \mid \hn}$ to achieve stability \cite{Mayne:2000} can be used as well. However, none of these embellishments will be used in this work.}
%Note that the predicted $\hat \vs_\tind$ depends on all previous $\vu_{0:K-1}$. 
%In our case, the prediction model $l$ is a neural network, i.e., the gradients with respect to the control signals can be efficiently computed by back-propagation. The cost function~\eqref{eq:MPC_cost_functionb} is then minimized with the BFGS algorithm.
%
When the control sequence $\vu_0^*,\dotsc, \vu_{K-1}^*$ is determined, the first control $\vu_0^*$ is applied to the system. After observing the next feature, NMPC repeats the entire optimization and turns the overall policy into a closed-loop (feedback) control strategy. 

Overall, we now have an online NMPC algorithm that, given a trained DDM, works indirectly on images by exploiting their feature representation. 

\subsection{Adaptive NMPC for Learning from Scratch}
%%%%%% Adaptive MPC 
We will now turn over to describe how adaptive NMPC can be used together with our DDM to address the pixels-to-torques problem and to learn from scratch.
% model is important -> MPC depends on it
At the core of our NMPC formulation lies the DDM, which is used to predict future features (and images) from a sequence of control signals. The quality of the NMPC controller is inherently bound to the prediction quality of the dynamical model, which is typical in model-based RL~\cite{Schneider1997,Schaal1997,DeisenrothFR:2015}.

%\begin{figure}
\begin{wrapfigure}{R}{0.6\textwidth}
\vspace{-6mm}
\begin{minipage}{0.6\textwidth}
\begin{algorithm}[H]
\begin{algorithmic} 
\STATE Follow a random control strategy and record data
\LOOP
  \STATE Update DDM with all data collected so far %using \eqref{eq:joint_training2}.
  \FOR{$t=0$ to $N-1$}
  \STATE Get current feature $\vz_\tind$ via the encoder
  \STATE $\vu_\tind^*\leftarrow \epsilon$-greedy NMPC policy using DDM pred.
%     \STATE $\gamma \sim \mathcal U[0,1]$
%     \IF{$\gamma > \epsilon$}
%           \STATE ${u}^*_\tind\leftarrow u^0_\tind$ (MPC solution)
%     \ELSE 
%         \STATE ${u}^*_\tind \sim \N(0,d^2)$
%     \ENDIF 
    \STATE Apply $\vu^*_\tind$ and record data
  \ENDFOR
\ENDLOOP
\end{algorithmic}
\caption{Adaptive online NMPC in feature space}
\label{alg:proposed}
%\end{figure}
\end{algorithm}
\vspace{-5mm}
\end{minipage}
\end{wrapfigure}

% iterative learning
To learn models and controllers from scratch, we apply a control scheme that allows us to update the DDM as new data arrives. In particular, we use the NMPC controller in an adaptive fashion to gradually improve the model by collected data in the feedback loop without any specific prior knowledge of the system at hand.
Data collection is performed in closed-loop (online NMPC), and it is divided into multiple sequential trials. After each trial, we add the data of the most recent trajectory to the data set, and the model is re-trained using all data that has been collected so far. 
% To account for previously collected data, we adapt the optimization problem \eqref{eq:joint_training} such that the prediction error term $V_{\text{P}}$ is the average of all prediction error terms from all trials and that the reconstruction error term $V_{\text{R}}$ uses all data collected so far.

Simply applying the NMPC controller based on a randomly initialized model would make the closed-loop system very likely to converge to a point, which is far away from the desired reference value, due to the poor model that cannot extrapolate well to unseen states. This would in turn imply that no data is collected in unexplored regions, including the region that we are interested in. There are two solutions to this problem: either we use a probabilistic dynamical model~\cite{Schneider1997,DeisenrothFR:2015} to explicitly account for model uncertainty and the implied natural exploration, or we follow an explicit exploration strategy to ensure proper excitation of the system. In this paper, we follow the latter approach. In particular, we choose an $\epsilon$-greedy exploration strategy where the optimal feedback $\vu_0^*$ at each time step is selected with a probability~$1-\epsilon$, and a random action is selected with probability~$\epsilon$.  

Algorithm~\ref{alg:proposed} summarizes our adaptive online NMPC scheme. We initialize the DDM with a random trial. We use the learned DDM to find an $\epsilon$-greedy policy using predicted features within NMPC. This happens online while the collected data is added to the data set, and the DDM is updated after each trial.

\section{Experimental Evaluation}
\label{sec:experimental}

In this section, we empirically assess the components of our proposed methodology for autonomous learning from high-dimensional synthetic image data, on learning the underlying dynamics of a single and a  planar double pendulum. The main lines of the evaluation are: (a) the quality of the learned DDM and (b) the overall learning framework.

% setting
In both experiments, we consider the following setting: We take screenshots of a simulated pendulum system at a sampling frequency of \SI{0.2}{s}. Each pixel $\vx_\tind^{(i)}$ is a component of the measurement $\vx_\tind\in\real^{N_\vx}$ and takes a continuous gray-value in the interval $[0,1]$. The control signals  $\vu_\tind$ are the torques applied to the system. No access to the underlying dynamics nor the state (angles and angular velocities) was available, i.e., we are dealing with a high-dimensional continuous time series. The challenge was to data-efficiently learn (a) a good dynamical model and (b) a good controller from pixel information only.
%We used a sampling frequency of \SI{0.2}{s} and a time horizon of \SI{200}{s} which corresponds to $1000$ frames per trial. \todo[inline]{didn't we use 100 sample for single and 1000 for double?}

% PCA + network architectures
To speed up the training process, we applied PCA prior to model learning as a pre-processing step to reduce the dimensionality of the original problem.
% The input dimension was reduced to $\text{dim}(\vx_\tind) = 100$ and $512$, for the single and the double planar pendulum respectively, prior to model learning using PCA.
% This step was chosen to speed up the training process. 
With these inputs, a $3$-layer auto-encoder was employed, such that the dimensionality of the features is optimal to model the periodic angle of the pendulums. The features $\vz_{\tind-1}, \vz_{\tind}$ and $\vu_{\tind}$ were later passed to the $3$-layer predictive feedforward neural network generating $\hat \vz_{\tind+1}$.
% Parameters
Furthermore, during training, the $\alpha$ parameter for encouraging consistent latent space predictions was set to $1$ for both experiments. While, in the adaptive NMPC, the $\lambda$ tuning parameter that penalizes the amplitude of the control signals, was set to $0.01$.

%\todo[inline]{We should somewhere say that the torque is our control input}

\subsection{Planar Pendulum}

\begin{figure}[b]
    \centering
    \figspace
    \begin{subfigure}[t]{0.48\textwidth}
    \includegraphics[width=\textwidth]{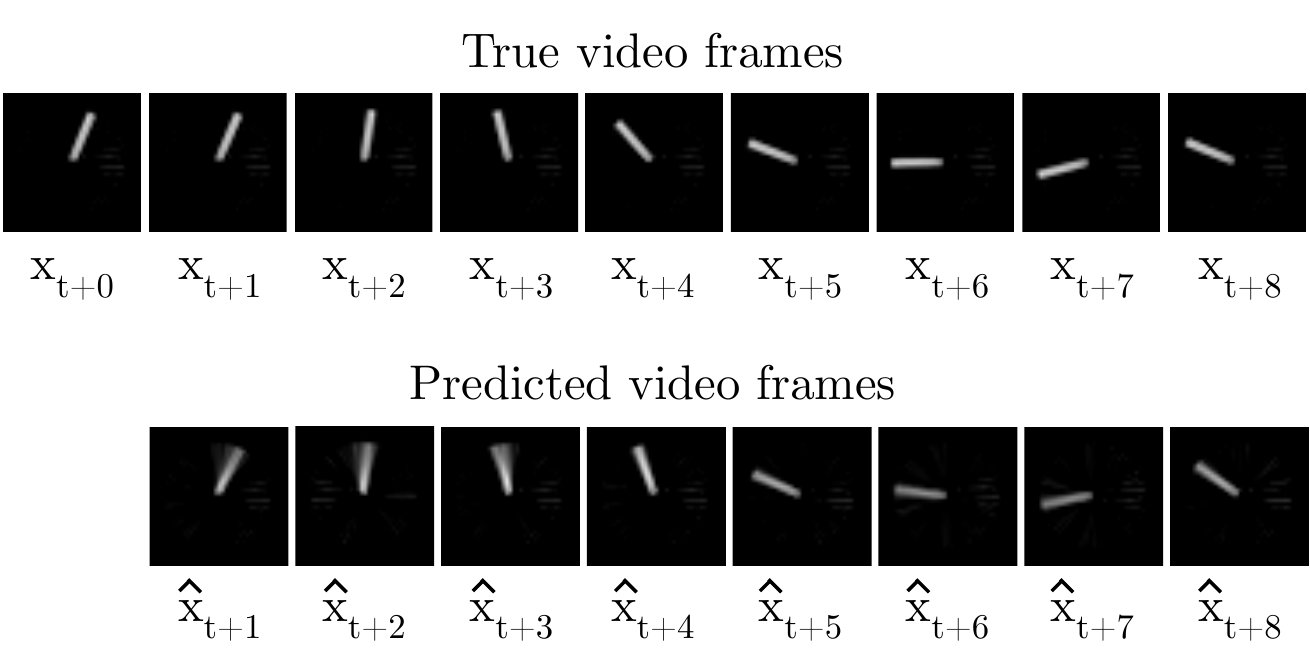}
    \caption{Planar single pendulum}
    \end{subfigure}%
\hfill
    \begin{subfigure}[t]{0.48\textwidth}
    \includegraphics[width=\textwidth]{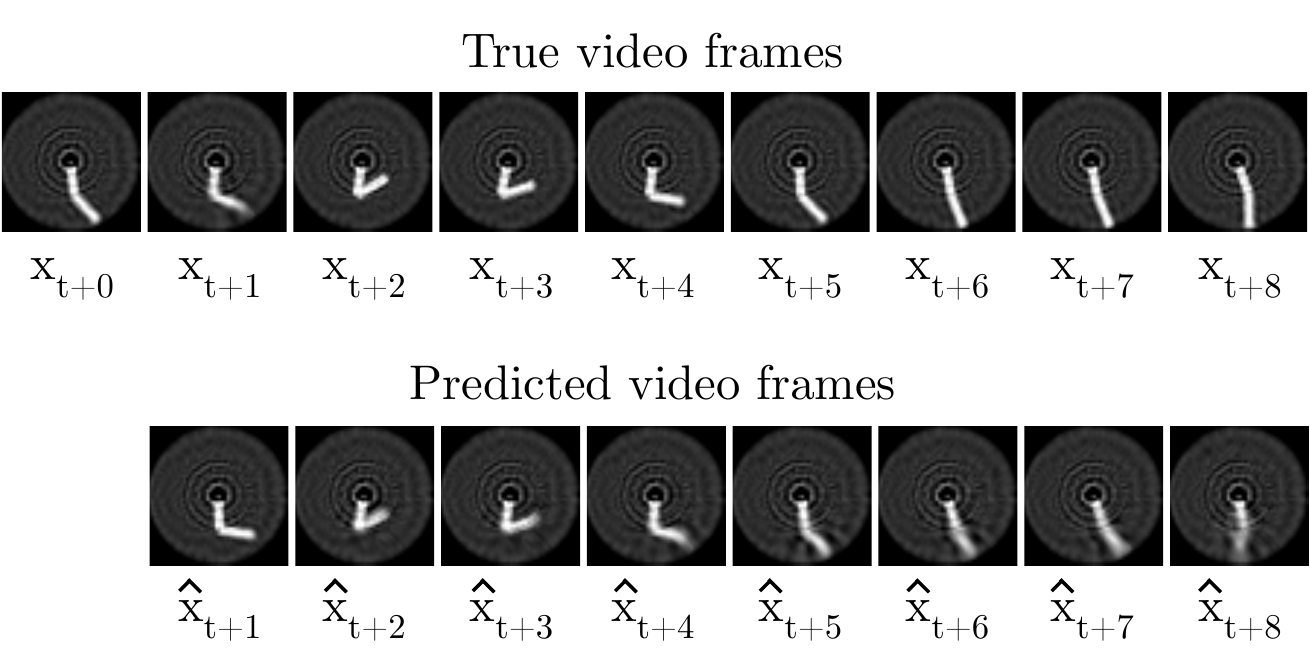}
    \caption{Planar double pendulum}
    \end{subfigure}%
    \caption{Long-term (up to eight steps) predictive performance of the DDM  controlling a planar pendulum (a) and a planar double pendulum (b): true (upper plot) and predicted (lower plot) video frames on test data.}
    \label{fig:pred_single_double}
\end{figure}

The first experiment evaluates the performance of the DDM on a planar pendulum, assembled by 1-link robot arm with length \SI{1}{m}, weight \SI{1}{kg} and friction coefficient \SI[per-mode=symbol]{1}{\newton\second\metre\per\radian}.%-1\,Nsm/rad.
The screenshots consist of $40\times40 = 1600$ pixels, and the input dimension has been reduced to $\text{dim}(\vx_\tind) = 100$ using PCA.
% network architecture
These inputs are processed by an encoder $f_{\text{enc}}$ with architecture: $100\times50$ -- ReLU -- $50\times50$ -- ReLU -- $50\times2$.

The low-dimensional features are of $\text{dim}(\vz_\tind) = 2$ in order to model the periodic angle of the  pendulum. To capture the \emph{dynamic} properties, such as angular velocity,  we concatenate two consecutive features $\vz_{\tind-1},\vz_{\tind}$ with the control signal $\vu_{\tind}\in\real^1$ and pass them through the predictive model $f_{\text{pred}}$, with architecture: $5\times100$ -- ReLU -- $100\times100$ -- ReLU -- $100\times2$. 
Note that the dimensionality of the first layer is given by $\text{dim}(\vz_{\tind-1})+\text{dim}(\vz_\tind)+\text{dim}(\vu_\tind)=5$.
Finally, the predicted feature $\hat \vz_{\tind+1}$, can be mapped back to $\hat \vx_{\tind+1}$ using our decoder, with architecture: $2\times50$ -- ReLU -- $50\times50$ -- ReLU -- $50\times100$.

\begin{wrapfigure}{l}{0.4\textwidth} 
    \centering
    \vspace{-5mm}
    \includegraphics[width=\linewidth]{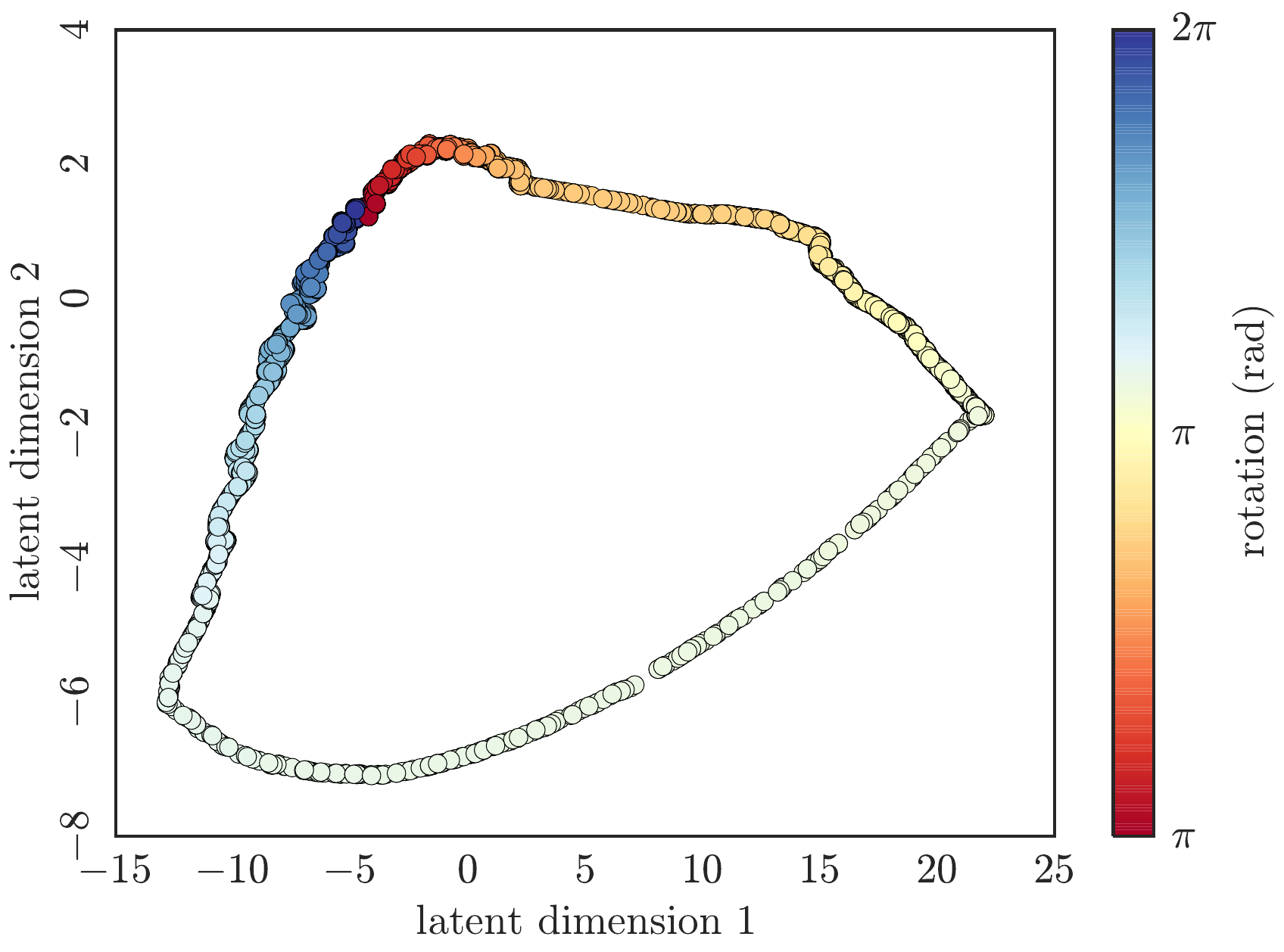}
    \caption{The feature space $\vz \in \real^2$ of the single pendulum experiment for different pendulum angles between $0^\circ$ and $360^\circ$, generated by the DDM.}
    \label{fig:latent_space}
%    \figspace
	\vspace{-5mm}
\end{wrapfigure}

The performance of the DDM is illustrated in Figure~\ref{fig:pred_single_double} on a test data set. The top row shows the true images and the bottom row shows the DDM's long-term predictions. 
%The model predicts future frames of the tile with high accuracy both for 1-step ahead and multiple steps ahead. 
The model yields a good predictive performance for both one-step ahead prediction and multiple-step ahead prediction, a consequence of (a) jointly learning predictor and auto-encoder, (b) concatenating features instead of images to model the dynamic behavior.

In Figure~\ref{fig:latent_space}, we show the learned feature space $\vz \in \real^2$ for different pendulum angles between $0^\circ$ and $360^\circ$. The DDM has learned to generate features that represent the angle of the pendulum, as they are mapped to a circle-like shape accounting for the wrap-around property of an angle. %suitable for modeling the dynamics of the pendulum. % However, the true state features were never available to the DDM. \todo[inline]{So i don't think we can say that there exist true features here}

%
%\begin{wrapfigure}{r}{0.4\textwidth}
%	\vspace{-4mm}
%    \centering
%    \includegraphics[width=\linewidth]{artwork/results_single2.pdf}
%    \caption{Results on learning a policy from scratch that moves the pendulum from downward ($\varphi = 0^\circ$) to upward ($\pm 180^\circ$). The horizontal axis shows the learning stages and the corresponding image frames available to the learner. The vertical axis shows the absolute error from the target state, averaged over the last 10 time steps of each test trajectory. The dashed line shows a \SI{10}{\degree} error, which indicates a ``good'' solution.}
%    \label{fig:ctrl_single}
%    \figspace
%\end{wrapfigure}

Finally, in Figure~\ref{fig:ctrl_single_double}, we report results on learning a policy that moves the pendulum from a start position $\varphi = 0^\circ$ to an upright target position $\varphi = \pm 180^\circ$. The reference signal was the screenshot of the pendulum in the target position. For the NMPC controller, we used a planning horizon of $K=15$ steps and a control penalty $\lambda=0.01$. For the $\epsilon$-greedy exploration strategy we used $\epsilon = 0.2$.

Figure~\ref{fig:ctrl_single_double} shows the learning stages of the system, i.e., the $18$ different trials of the NMPC controller.
Starting with a randomly initialized model, $500$ images were appended to the dataset in each trial. As it can be seen, starting already from the first controlled trial, the system managed to control the pendulum successfully and bring it to a position less than \SI{10}{\degree} from the target position. This means, the solution is found very data efficiently, especially when we consider that the problem is learned from pixels information without access to the ``true'' state.
%Note that in the later stages of learning (blue trajectories), the system was able to solve the task in fewer time steps.

\subsection{Planar Double Pendulum}

\begin{figure}[b]
	\figspace
    \centering
    \begin{subfigure}[t]{0.48\textwidth}
    \includegraphics[width=\textwidth]{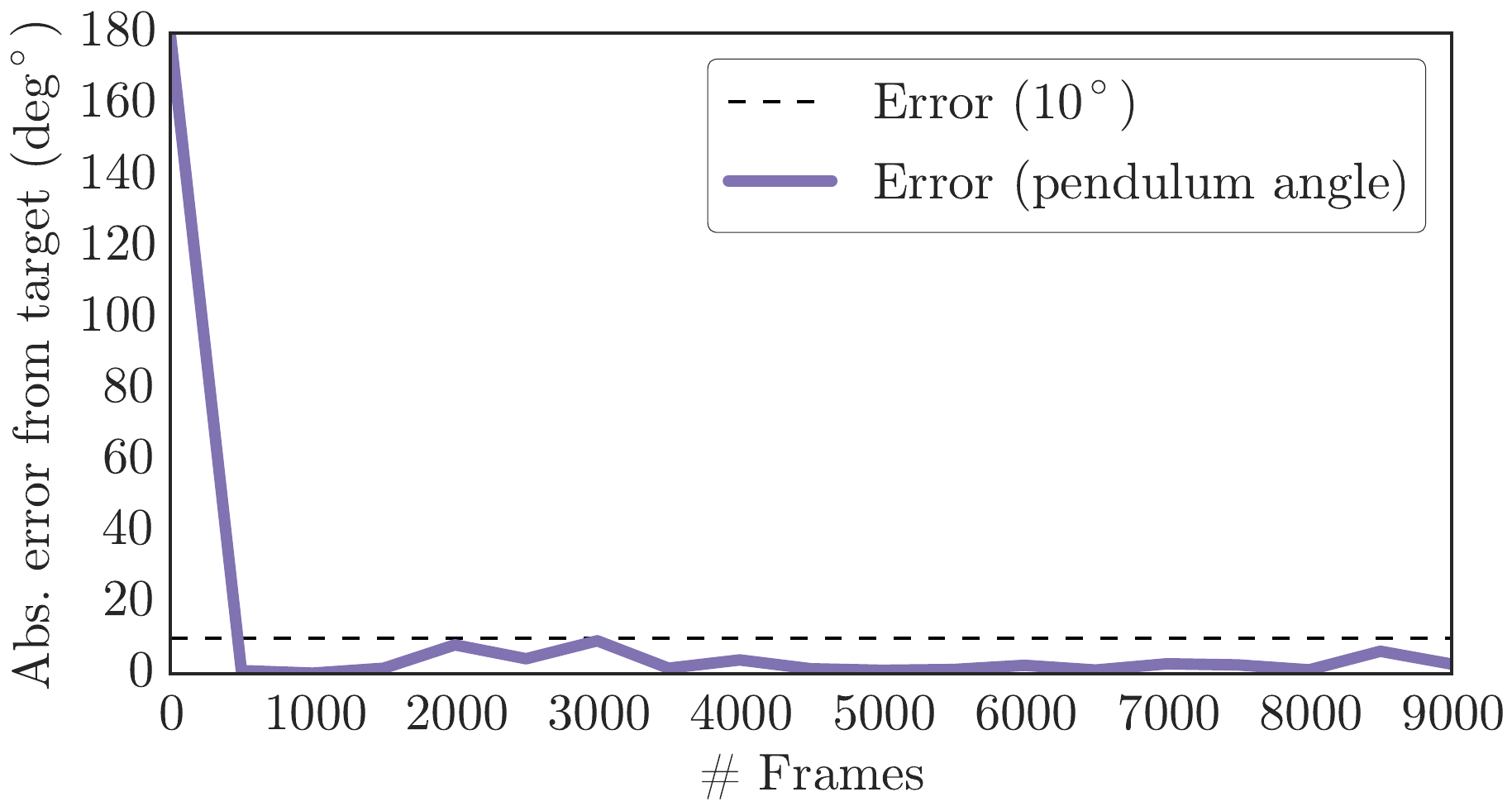}
    \caption{Planar single pendulum}
    \end{subfigure}%
    \hfill
    \begin{subfigure}[t]{0.48\textwidth}
    \includegraphics[width=\textwidth]{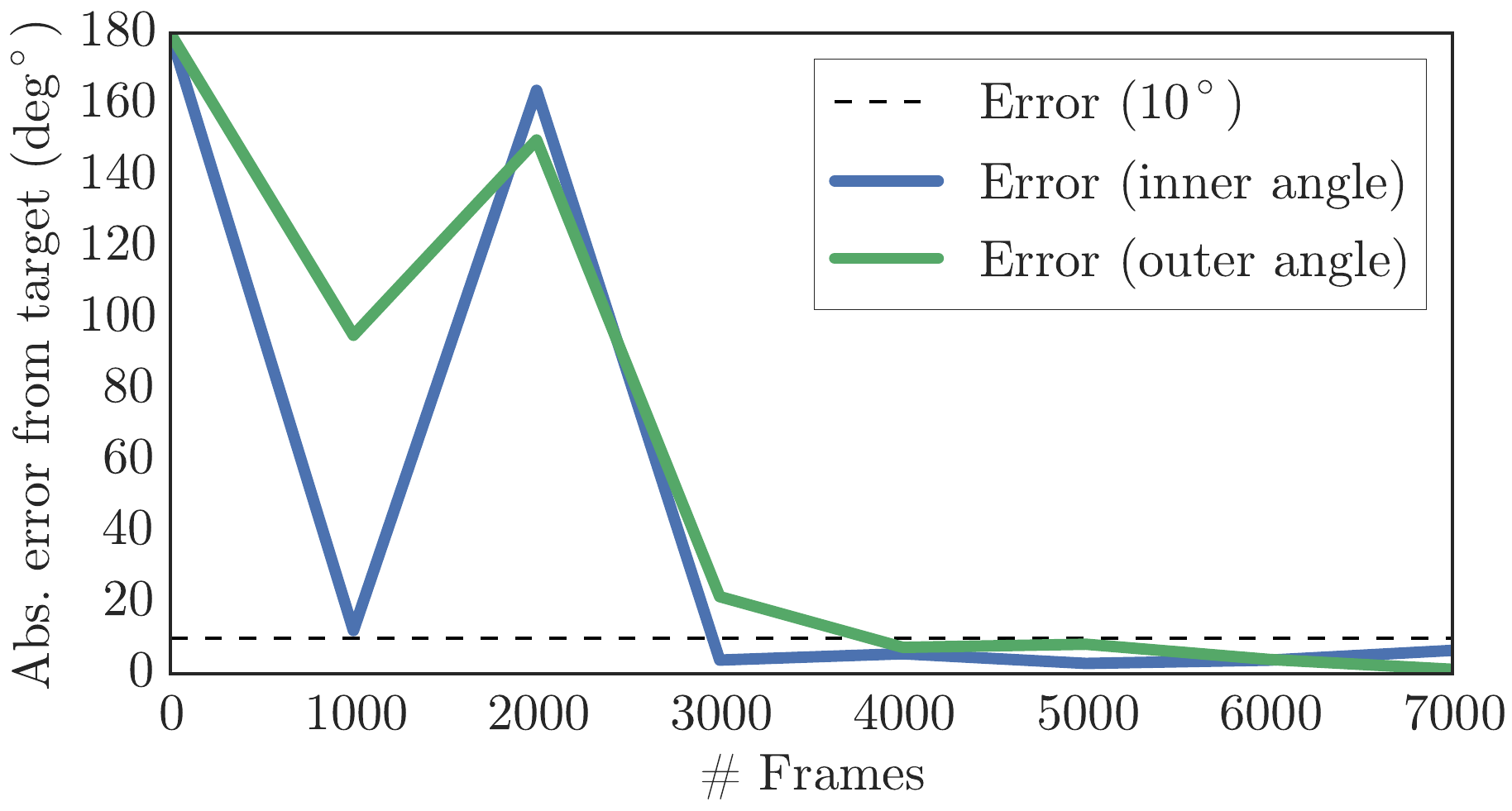}
    \caption{Planar double pendulum}
    \end{subfigure}
    \caption{Results on learning a policy that moves the single (a) and the double (b) pendulum systems from $\varphi = 0^\circ$ to $\varphi = \pm 180^\circ$, in $100$ time-steps. The horizontal axis shows the learning stages and the corresponding image frames available to the learner. The vertical axis shows the absolute error from the target state, averaged over the last 10 time steps of each test trajectory. The dashed line shows a \SI{10}{\degree} error, which indicates a ``good'' solution.}
    \label{fig:ctrl_single_double}
	\figspace
\end{figure}

In this experiment, a  planar double pendulum is considered assembled by 2-link robot arm with length \si{1}{m} and \si{1}{m} respectively, weight \si{1}{kg} and \si{1}{kg} and friction coefficients \SI[per-mode=symbol]{1}{\newton\second\metre\per\radian}. Torques can be applied at both joints.
The screenshots consist of $48\times48=2304$ pixels, and the input dimension has been reduced to $\text{dim}(\vx_\tind) = 512$ prior to model learning using PCA to speed up the training process.
% network architecture
The encoder architecture is: $512\times256$ -- ReLU -- $256\times256$ -- ReLU -- $256\times4$, and the decoder vice versa. The low-dimensional embeddings $\text{dim}(\vz_\tind) = 4$ and the architecture of the predictive model  was: $10\times200$ -- ReLU -- $200\times200$ -- ReLU -- $200\times4$.

%\begin{wrapfigure}{l}{0.4\textwidth}
%	\vspace{-5mm}
%    \centering
%    \includegraphics[width=\linewidth]{artwork/pred_double.pdf}
%    \caption{Long-term (up to eight steps) predictive performance of the DDM controlling a double pendulum: True and predicted video frames on test data.}
%    \label{fig:pred_double}
%    \figspace
%\end{wrapfigure}

The predictive performance of the DDM is shown in Figure~\ref{fig:pred_single_double} on a test data set. The performance of the controller is depicted in Figure~\ref{fig:ctrl_single_double}. We used $7$ trials with the downward initial position $\varphi = 0^\circ$ and upward target $\varphi = \pm 180^\circ$ for the angle of both inner and outer pendulums. The figure shows the error after each trial (1000 frames) and clearly indicates that after three controlled trials a good solution is found, which brings both pendulums within a  $10^\circ$ range to the target angles.

%that this time had $1,000$ additional points. 
Despite the high complexity of the dynamical system, our learning framework  manages to successfully control both pendulums after the third trial in nearly all cases.

% \begin{itemize}
% \item Planar double pendulum (this is the minimum)
% \item Comparison with PILCO (Marc)
% \item Comparison with Niklas' PCA-based AE approach (if we choose to go for the convolutions)
% \item Comparison with E2C
% \item Show training time, performance, number of parameters
% \item If possible: show some modeling pictures of the 3D robot arm (John), but only if they look great.
% \end{itemize}

\subsection{Comparison with State-of-the-Art}

The same experiments were executed employing PILCO~\cite{Deisenroth2011c}, a state of the art policy search method, under the following settings: (a) PILCO has access to the true state, i.e., the angle $\varphi$ and angular velocity $\dot \varphi$; (b) A deep auto-encoder is used to learn two-dimensional features $\vz_t$ from images, which are used by PILCO for policy learning.
In the first setting (a) PILCO managed to successfully reach the target after the second and the third trial in the two experiments,  respectively. However,  in setting (b), PILCO did not manage to learn anything meaningful at all. The reason why PILCO could not learn on auto-encoder features is that these features were only trained to minimize the reconstruction error. However, the auto-encoder did not attempt to map similar images to similar features, which led to zig-zagging around in feature space (instead of following a smooth manifold as in Figure~\ref{fig:latent_space}), making the model learning part in feature space incredibly hard~\cite{wahlstrom2015learning}. 

We modeled and controlled equally complex models with E2C~\cite{watter2015e2c}, but at the same time our DDM requires ${\approx}4\times$ fewer neural network parameters if we use the same PCA pre-processing step within E2C. The reason lies in our efficient processing of the dynamics of the model in the feature space instead of the image space. This number increases up to ${\approx}20\times$ fewer parameters than E2C without the PCA pre-processing step.

The number of parameters can be directly translated to reduced training time, and increased data efficiency. Employing the adaptive model predictive control, our proposed DDM model requires significantly less data samples, as it efficiently focuses on learning the latent space towards the reference target state. Furthermore, the control performance of our model is gradually improved in respect to the number of trials. As proved by our experimental evaluation we can successfully control a complex dynamical system, such as the  planar double pendulum, with less than $4\thinspace 000$ samples. This adaptive learning approach can be essential in problems with time and hardware constraints. 

%More specifically, in the single planar pendulum case under the same training data and settings, we used $26,404$ parameters, when following the suggested E2C architecture would require ${\sim}6,400,000$ or ${\sim}550,000$ if E2C-Conv with convolutional layers was used. Similarly for the double planar pendulum, our approach used ${\sim}440,008$ parameters, when E2C and E2C-Conv would require ${\sim}19,900,000$ and ${\sim}1,550,000$ respectively. % \todo[inline]{But E2C used gravity, right? Is this comparison fair?}
%The number of parameters can be directly translated to reduced training time, and increased data efficiency. % More specifically, we showed that we can control a single pendulum with $1,000$ samples, when E2C requires $3,000$ for a much simpler experiment.

\section{Conclusion}

We proposed a data-efficient model-based RL algorithm that learns closed-loop policies in continuous state and action spaces directly from pixel information. The key components of our solution are (a) a deep dynamical model (DDM) that is used for long-term predictions via a compact feature space, (b) a novel training objective that encourages consistency in the latent space, paving the way toward more accurate long-term predictions, and (c) an NMPC controller that uses the predictions of the DDM to determine optimal actions on the fly without the need for value function estimation. For the success of this RL algorithm it is crucial that the DDM learns the feature mapping and the predictive model in feature space jointly to capture dynamical behavior for high-quality long-term predictions. Compared to state-of-the-art RL our algorithm learns fairly quickly, scales to high-dimensional state spaces and facilitates learning from pixels to torques.

\subsubsection*{Acknoledgements}
We thank Roberto Calandra for valuable discussions in the early stages of the project. The Tesla~K40 used for this research was donated by the NVIDIA Corporation.

% open-source
%To facilitate future research, we will open source our implementation in %Torch7~\cite{collobert2011torch7} and Matlab.

%%% bibliography

\newpage
\subsubsection*{References}
\bibliography{references}

\end{document}

%% file: include/murphy.tex
\newcommand{\real}{\mathbb{R}}
\newcommand{\loss}{\calL}

\newcommand{\myvec}[1]{\mathbf{#1}}

\newcommand{\vs}{\myvec{s}}

\newcommand{\vu}{\myvec{u}}

\newcommand{\vx}{\myvec{x}}

\newcommand{\vz}{\myvec{z}}

\newcommand{\calD}{{\cal D}}

\newcommand{\calL}{{\cal L}}

\newcommand{\data}{\calD}

\newcommand{\be}{\begin{equation}}
\newcommand{\ee}{\end{equation}}
\newcommand{\bea}{\begin{eqnarray}}
\newcommand{\eea}{\end{eqnarray}}
\newcommand{\beaa}{\begin{eqnarray*}}
\newcommand{\eeaa}{\end{eqnarray*}}

\DeclareMathAlphabet{\mathpzc}{OT1}{pzc}{m}{n}